\documentclass{article}

\usepackage[preprint]{neurips_2026}  

\usepackage[utf8]{inputenc}
\usepackage[T1]{fontenc}
\usepackage{hyperref}
\hypersetup{hidelinks}
\usepackage{url}
\usepackage{booktabs}
\usepackage{amsmath}
\usepackage{amssymb}
\usepackage{graphicx}
\usepackage{subcaption}
\usepackage{multirow}
\usepackage{xcolor}
\usepackage{microtype}
\usepackage{enumitem}

\graphicspath{{figures/}}

\newcommand{\tta}{\ensuremath{\tau}}
\newcommand{\ci}[2]{\,[#1,#2]}
\newcommand{\FA}{\textrm{FA}}
\newcommand{\cov}{\textrm{cov}}

\title{Measuring the Microtask Eligibility Gap: When Is an Off-the-Shelf SLM Enough for an Agent Harness?}

\makeatletter
\newif\ifshowcredits
\if@anonymous
  \newcommand{\authornote}{}%
  \showcreditsfalse
\else
  \newcommand{\authornote}{\thanks{Corresponding author (\texttt{jundongh@alumni.upenn.edu}). Jundong Hu led and carried out the research end to end.}}%
  \showcreditstrue
\fi
\makeatother
\renewcommand{\acksection}{\section*{Acknowledgments}}
\author{%
  Jundong Hu\authornote \\
  PayPal AI \\
  \texttt{jundhu@paypal.com} \\
  \And
  Shekar Ramachandran \\
  PayPal AI \\
  \texttt{sheramachandran@paypal.com} \\
}

\begin{document}
\ifdefined\linenumbers\linenumbers\fi
\maketitle

\begin{abstract}
Agent harnesses increasingly want to run small language models (SLMs) on the microtasks
around a frontier large language model (LLM) planner: auto-approving shell commands, writing memory, selecting tools, ranking past
turns. We ask whether off-the-shelf SLMs meet practitioner-defined thresholds and, when they fail, why, and
whether quantization changes the answer. We build a benchmark of 4 such microtasks with fixed prompts and automatic metrics, each with a
pre-specified threshold \tta{} anchored to a cheap non-LLM baseline and a CI-aware
eligibility rule (a configuration passes only if its confidence bound clears \tta). Sweeping Qwen3
\{0.6, 1.7, 4, 8\}B at their best (FP16, greedy, one frozen prompt, no tuning), we find an
eligibility gap: 0 of 16 (4 tasks $\times$ 4 models) configurations pass (verified by checking the raw outputs and parser
behavior). A logprob decision-threshold diagnostic (T1/T3/T4; T2 via a context-length/cascade
probe) separates the failures into capability deficits and failures that can be addressed by changing the
decoding threshold (4 regimes). Quantization to 4-bit (RTN/GPTQ/AWQ) does damage that depends on model size
and moves no configuration into eligibility (certified on the
reconstructable hard-label tasks T1/T3, diagnostic/windowed robustness on T2/T4), so the gap tracks
model size more than precision; it replicates on Llama-3.x (12/12 ineligible) and is robust to the
anchor choice (a $\tau$-sweep) and to prompt wording (0/112 eligible across the original plus 3
neutral paraphrases per cell). The practical implication: place SLMs behind a baseline that
meets the CI-backed threshold, and use the SLM only where the baseline fails to meet the threshold; e.g.\ a 4B re-ranker over a BM25
shortlist beats BM25 ($+0.047\ci{0.020}{0.073}$, without itself certifying eligibility).
\end{abstract}

\section{Introduction}
An agent spends most of its tokens on microtask decisions in the harness rather than on the single
headline reasoning step: whether to auto-approve a shell command, write a fact to
memory, select a tool, or rank which of the last several hundred turns are relevant. These are exactly the
calls a deployer would want to hand to a small, cheap local model, and a growing line of work argues SLMs
are a good fit for such agentic subtasks~\citep{belcak2025smallagents}. A representative
production harness makes dozens per session, so the open question is empirical: can an off-the-shelf SLM
be trusted with them?

We treat ``trusted with'' operationally. For each microtask we set a pre-specified eligibility
threshold \tta, anchored to a cheap non-LLM baseline a practitioner already has (a deliberation
safety rule, BM25, a write-heuristic, a TF-IDF tool shortlist). An SLM is eligible only if it
clears \tta{} with its confidence interval (CI): it must exceed the non-LLM baseline under the stated rule.
Everything (test split, prompts, metrics, \tta) is frozen before evaluation. We evaluate one frozen prompt per microtask for the
headline; to rule out a wording artifact we separately re-test each cell under 3 neutral
paraphrases (4 phrasings per task, 112 configurations across both model families;
Appendix~\ref{app:paraphrase}).

Across Qwen3 \{0.6, 1.7, 4, 8\}B in their best case (FP16, greedy, one frozen prompt), none of
the 16 (task, model) configurations are eligible. This empty set is the central result: anchoring
each task to the baseline you would otherwise deploy turns the measurement into a direct basis for
choosing between the baseline and an SLM, and makes the eligibility verdict precise and CI-defensible. We isolate
\emph{capability eligibility} from run-time cost/latency/energy (hardware-dependent, left to systems work):
``eligible'' means ``clears the quality bar,'' checked before cost.

\paragraph{Contributions.}
\begin{enumerate}
  \item A \textbf{frozen 4-microtask agent-harness benchmark} (T1 shell auto-approval, T2 memory recall,
    T3 memory write-decision, T4 tool selection), judge-free and closed-set, each with a
    baseline-anchored \tta{} and CI-aware rule (§\ref{sec:setup}).
  \item \textbf{Result 1: the eligibility gap exists.} 0 of 16 FP16 configurations pass, verified against
    raw outputs (§\ref{sec:gap}).
  \item \textbf{Result 2: diagnosing the gap.} A logprob decision-threshold diagnostic sorts the failures
    into 4 regimes: capability deficit, decoding/operating-point repair, capacity repair, and
    architecture repair (§\ref{sec:decomp}).
  \item \textbf{Result 3: baseline-gated cascades.} A measured non-LLM$\rightarrow$SLM
    crossover (cf.\ FrugalGPT): a BM25 shortlist$\rightarrow$4B re-rank
    reverses T2, and retrieval gating marks the T4 calls that need no model (§\ref{sec:cascade}).
  \item \textbf{Result 4: quantization damage depends on model size and makes no configuration eligible.} 4-bit RTN/GPTQ/AWQ
    move none into eligibility (certified on hard-label T1/T3$\ge$1.7B, surrogate-supported on T2/T4), so
    the gap tracks model size more than precision (§\ref{sec:quant}).
\end{enumerate}

\section{Related Work}
\textbf{SLM-as-agent advocacy}~\citep{belcak2025smallagents} argues most agent calls are narrow and
SLM-suitable. We operationalize and test that thesis with a measured, baseline-anchored
eligibility rule on real harness microtasks, and find the gap.
\textbf{Agent/tool benchmarks}~\citep{liu2024agentbench,qin2024toolllm,bfcl2025} evaluate agents across
interactive environments; they do not anchor to a non-LLM baseline, use a CI-aware eligibility rule, cover
shell safety / memory microtasks, or decompose capability vs.\ calibration. We reuse Berkeley Function Calling Leaderboard (BFCL) data for T4 but recast
it as an eligibility test.
\textbf{Calibration / logit-signal work}~\citep{guo2017calibration} shows a decision signal can be present
while decoding is miscalibrated; we apply this lens to separate capability deficit from calibration
failure on agent decisions and tie it to deployability (calib$\rightarrow$test: a threshold selected on the calibration split rather than the test split).
\textbf{Classical retrieval / re-ranking}~\citep{robertson2009bm25,nogueira2019rerank} motivates our
cascade design; the finding is the case in which the SLM exceeds BM25: a 4B re-ranker beats BM25 over a shortlist
while full-index ranking fails.
\textbf{Model routing / cascades}~\citep{chen2024frugalgpt} learn cost-aware LLM$\rightarrow$LLM
escalation; our contribution is the measured non-LLM$\rightarrow$SLM crossover after a
baseline-anchored eligibility test, rather than the cascade mechanism itself.
\textbf{Harness adaptation for SLM agents}~\citep{yang2026betterharnesses} adapts contexts/tools/loops
around SLMs to close the SLM--frontier gap; we ask the complementary, pre-adaptation question: under
a frozen harness, does the off-the-shelf model already meet the non-LLM baseline, and why it fails.
\textbf{Post-training quantization}~\citep{frantar2023gptq,lin2024awq,dettmers2023case4bit} is usually
measured by perplexity/general benchmarks; we measure it on task eligibility for agent microtasks
and show damage depends on model size and never flips eligibility.
Our T2/T3 data are adapted from long-term conversational memory~\citep{maharana2024locomo}; models are
Qwen3~\citep{qwen3} and, for cross-family replication, Llama-3.x~\citep{llama3}.

\section{Experimental Setup}\label{sec:setup}
We study 4 microtasks, one frozen prompt each and no per-model tuning (Table~\ref{tab:setup}); the same 4 (T1--T4) are the unit of analysis for every result that follows. Each has a
primary metric and a threshold \tta{} anchored to a cheap non-LLM baseline a deployer already has.
Eligibility is CI-aware: a configuration passes only if its confidence bound clears \tta{}: T1
Wilson~\citep{wilson1927probable} ($\FA$-upper $\le\tta$ \emph{and} $\cov$-lower $\ge\tta$), T2/T3
10k-bootstrap~\citep{efron1979bootstrap} lower $\ge\tta$, T4 Wilson lower $\ge\tta$; beating \tta{} on the
point estimate is not enough. A second classical baseline per task lands within a few points of each \tta{} (for T1, a composite safety/coverage frontier),
and the verdict survives a $\tta$-sensitivity sweep (Appendix~\ref{app:anchor}). Anchors are frozen on the
\emph{calibration} split: the T2 bar $\tta{=}0.479$ is BM25 recall@5 on calib vs.\ $0.440$ on the
$n{=}984$ test set (the cascade comparator), so the frozen bar is if anything slightly stricter than
test-set BM25.

\begin{table}[t]
\centering\small
\caption{Benchmark setup. One prompt per task; \tta{} anchored to a non-LLM baseline. Test sizes:
T1 301 / T2 984 / T3 3032 / T4 600.}
\label{tab:setup}
\resizebox{\textwidth}{!}{%
\begin{tabular}{@{}llll@{}}
\toprule
Task & Decision & Primary metric & \tta{} (frozen) / anchor \\
\midrule
T1 & safe shell auto-approve & false-approval $\FA\!\downarrow$ \& coverage $\cov\!\uparrow$ & $\FA\le0.167$ \& $\cov\ge0.921$ / deliberation+regex \\
T2 & memory relevance recall & Recall@5 & $\ge0.479$ / BM25 Recall@5 \\
T3 & memory write / no-write & write-F1 & $\ge0.656$ / write-heuristic \\
T4 & tool selection & correct-tool acc (callable) & $\ge0.913$ / TF-IDF shortlist top-1 \\
\bottomrule
\end{tabular}%
}
\end{table}

\textbf{Ladder \& inference.} Qwen3 dense non-thinking \{0.6, 1.7, 4, 8\}B, greedy,
\texttt{enable\_thinking=False}, one frozen prompt per task; all scale claims are within-family.
\textbf{Quantization axis:} 12 int4 checkpoints ($=$ 4 sizes $\times$ \{RTN (round-to-nearest), GPTQ, AWQ\}), packed
compressed-tensors, group-128 symmetric ($\sim$4.125 effective bits/weight), evaluated on the identical
frozen protocol. \textbf{Diagnostic:} we also record the model's continuation-logprob margin per item
(T1: $\log P(\texttt{AUTO\_APPROVE})-\log P(\texttt{ESCALATE})$; T3: $\log P(\texttt{\{})-\log
P(\texttt{NONE})$; T4: $\log P(\texttt{NO\_CALL})-\log P(\texttt{\{})$) and report the threshold-free area under the ROC curve (AUROC),
calib$\rightarrow$test (threshold on calib, evaluate once on test: deployable), and
oracle-on-test (threshold on test: a reachability ceiling, never a deployable claim). Test split,
prompts, \tta, metrics, and the eligibility rule were fixed before any evaluation. Freezing the harness is
part of the estimand and does not evaluate whether adaptation would help: we measure off-the-shelf eligibility at an
existing decision point before any task- or model-specific harness
engineering~\citep{yang2026betterharnesses}.

\section{Result 1: The Eligibility Gap Exists}\label{sec:gap}
In the models' best case, every (task, model) configuration is ineligible, most by substantial margins
(Table~\ref{tab:elig}). The failures take 2 forms. T1 (safety) shows bimodal policy collapse:
0.6B escalates almost everything ($\cov 0.029$), 1.7B approves almost everything ($\FA 0.988$), and 4B/8B
reach high coverage but remain well above the false-approval threshold ($\FA 0.414/0.463 \gg 0.167$), so none hits $\FA\le0.167 \wedge
\cov\ge0.921$. T2 (recall) rises with scale then dips (4B $0.350>$8B $0.292$), staying below lexical BM25 ($0.479$).
T3 (write) is a steady increase with model size; 8B F1 $0.634$ is closest but short of $0.656$.
T4 (tool) is an abstention/precision tradeoff: 0.6B's high correct-tool ($0.900$) is a never-abstain
artifact (irrelevance-acc $0.247$); the best-performing selector is 8B, but it reaches only $0.784\ll0.913$. No CI
bound reaches \tta{} (closest lower bounds T2 $0.322$, T3 $0.613$, T4 $0.869$, the 0.6B never-abstain cell;
the best-performing 8B selector's is $0.746$; T1 8B $\FA$-ub $0.540$).

\begin{table}[t]
\centering\small
\caption{FP16 eligibility sweep (point estimates). All 16 configurations are ineligible: no
CI bound clears \tta{} (bounds in text). T1 requires both $\FA\!\le\!0.167$ and $\cov\!\ge\!0.921$.}
\label{tab:elig}
\begin{tabular}{@{}lcccc@{}}
\toprule
 & T1 $\FA\!\downarrow$ \& $\cov\!\uparrow$ & T2 R@5 & T3 wF1 & T4 tool \\
\midrule
Qwen3-0.6B & 0.012, 0.029 & 0.041 & 0.557 & 0.900 \\
Qwen3-1.7B & 0.988, 1.000 & 0.125 & 0.579 & 0.758 \\
Qwen3-4B   & 0.414, 0.971 & 0.350 & 0.598 & 0.707 \\
Qwen3-8B   & 0.463, 0.993 & 0.292 & 0.634 & 0.784 \\
\midrule
\tta{}     & $\le$0.167 \& $\ge$0.921 & $\ge$0.479 & $\ge$0.656 & $\ge$0.913 \\
\bottomrule
\end{tabular}
\end{table}

\textbf{Raw-output spot-checks} ($n\!=\!40$/cell) confirm the reported behaviors: T1@0.6B escalates 39/40;
T1@1.7B auto-approves 40/40; T4@0.6B emits valid tool-call JSON with 0/40 abstentions; T2@4B produces
well-formed multi-session ID lists with imperfect ranking. The empty eligibility set comes from model behavior; parser checks do not change it.

\textbf{Anchor-robust.} Sweeping \tta{} over 28 FP cells with no re-run
(Appendix~\ref{app:anchor}), the empty-eligible set holds over the full tested range: the first
single-cell flip needs \tta{} to fall to $0.322$ (T2) or $0.613$ (T3), or a $2.9\times$ T1 safety
relaxation (the T4 first flip at $\tta'{=}0.869$ is 0.6B's never-abstain inflation, an argument to
keep $0.913$); the gap reflects the models rather than the threshold. \textbf{Cross-family
replication.} Re-running the full protocol on a Llama-3.x \{1,3,8\}B ladder (FP16, same frozen
prompts/\tta/splits, no tuning) reproduces the gap (12/12 ineligible), the failure modes, and the
four-regime decomposition (Appendix~\ref{app:xfam}); the qualitative pattern and eligibility verdict transfer, though not the exact
magnitudes.

\section{Result 2: Diagnosing the Gap}\label{sec:decomp}
A hard greedy label cannot tell us whether the signal to decide correctly exists. Using the logprob
margin (no change to the frozen generation path), we compute threshold-free AUROC (is the signal present?)
and calib$\rightarrow$test / oracle thresholded metrics (Fig.~\ref{fig:diag}). This separates
capability deficit (signal absent) from decoding/operating-point-recoverable failure
(signal present, greedy decoding fails to use it), yielding 4 regimes:

\begin{figure}[t]
\centering
\begin{subfigure}{0.32\textwidth}\includegraphics[width=\linewidth]{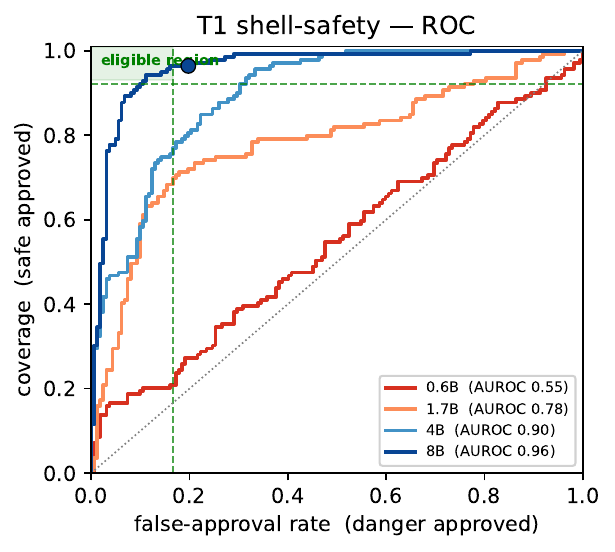}\caption{T1 ROC}\end{subfigure}\hfill
\begin{subfigure}{0.32\textwidth}\includegraphics[width=\linewidth]{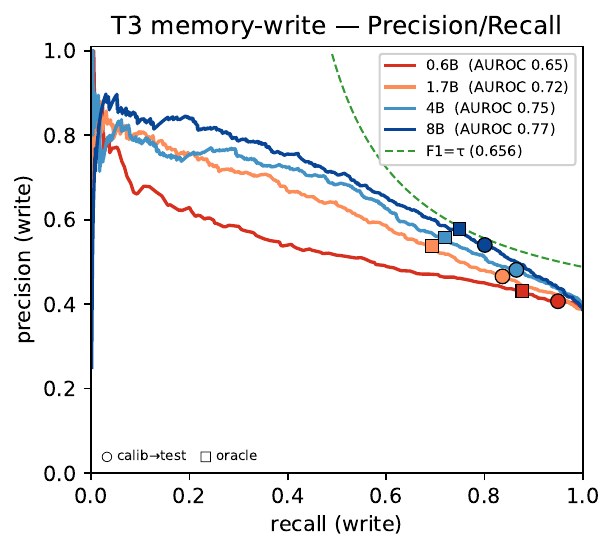}\caption{T3 P--R}\end{subfigure}\hfill
\begin{subfigure}{0.32\textwidth}\includegraphics[width=\linewidth]{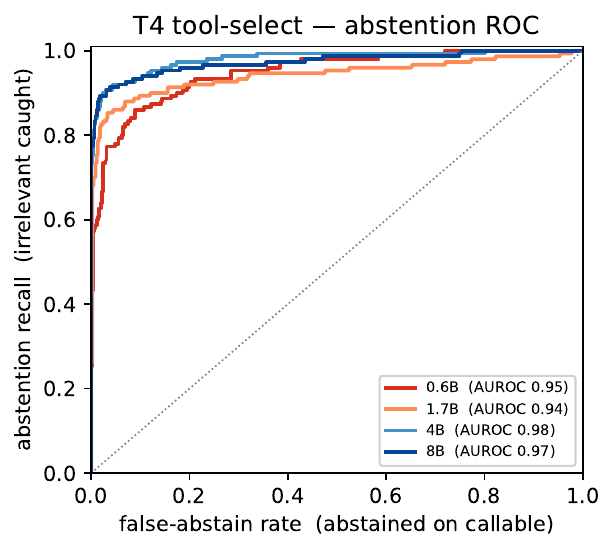}\caption{T4 abstention ROC}\end{subfigure}
\caption{Decision-threshold diagnostic (FP16). T1: only the 8B ROC enters the eligibility region. T3: all
operating points, including the oracle, sit below the $\text{F1}=0.656$ iso-contour. T4: abstention is
separable (AUROC 0.94--0.98) at every scale, a decoding artifact rather than missing capability.}
\label{fig:diag}
\end{figure}

\begin{enumerate}
  \item \textbf{Decoding/operating-point recoverable}: strong signal, greedy decoding fails to use it, a threshold
    recovers it, though the reason differs by cell. T4 abstention is a pure decision-boundary
    mismatch on a well-calibrated signal (AUROC $0.942$--$0.980$, expected calibration error [ECE] $0.05$--$0.07$; a threshold
    recovers abstain-accuracy to 0.87--0.92 while still calling on 89--97\% of callable items, so the ``0.6B
    never abstains'' degeneracy is entirely decoding), whereas T1@8B is genuine
    overconfidence (AUROC 0.959, ECE $0.256$; oracle-on-test just reaches the eligibility region
    at $\FA 0.167$, a ceiling, while honest calib$\rightarrow$test pulls greedy $\FA 0.463\to0.198$ but
    $\FA$-ub $0.266$ stays CI-ineligible).
  \item \textbf{Capacity gap, calibration insufficient}: moderate signal; even the oracle falls short.
    T3 AUROC $0.650\to0.771$; 8B greedy F1 $0.634\to$ calib$\rightarrow$test $0.645\to$
    oracle-on-test $0.652$, still below \tta{} 0.656 by 0.004.
  \item \textbf{Long-context-capacity deficit, architecture-fixable}: T2 full-index fails, but a cascade
    beats BM25 at 4B without reaching eligibility (§\ref{sec:cascade}).
  \item \textbf{Pure capability deficit}: no signal; T1@0.6B AUROC 0.548 (on the ROC diagonal).
\end{enumerate}
This turns the aggregate failure result into a practical diagnosis: decoding, a larger model, or a different architecture. The diagnostic is post-hoc; the frozen greedy
result stays primary, and we report calib$\rightarrow$test (deployable), oracle only as a reachability
ceiling. The 4 regimes form a qualitative spectrum rather than sharp cutoffs: at these per-task $n$ (T1
as small as 301) we do not claim adjacent-regime AUROC gaps are statistically distinguishable (bootstrap
CIs, both families, Table~\ref{tab:aurocci}), and label each cell by its main observed pattern. A reliability
check corroborates this: the decision logprobs are overconfident on T1/T3 (ECE $0.24$--$0.53$) yet
well-calibrated on T4 abstention (ECE $0.05$--$0.07$), so T4's ``never abstains'' is an
operating-point choice rather than a mis-estimated probability (Appendix~\ref{app:ece}).

\section{Result 3: Baseline-Gated Cascades}\label{sec:cascade}
\textbf{T4: retrieval-gated cascade.} Tool selection is regime-dependent: single/few-tool catalogues give
TF-IDF $\approx1.00$ ($n\!=\!75$/subset, Wilson $\ci{0.951}{1.000}$; no model needed), while realistic
multi-tool catalogues (BFCL \texttt{live\_multiple}, $\sim$4 similar tools) drop retrieval to $0.587$
($\ci{0.474}{0.691}$; non-overlapping), so a model is needed here. Retrieval cannot construct
arguments or abstain at all. A practical design: retrieval handles confident selection at zero model
cost, escalating to an SLM only for argument construction, abstention, and ambiguous multi-tool selection.

\textbf{T2: BM25$\rightarrow$SLM re-rank cascade.} Full-index T2 conflates long-context ranking capacity
with relevance judgment. In a new labeled condition, BM25 shortlists the 419-turn index to top-20 and the
SLM re-ranks within it under the same frozen prompt (BM25 R@20 ceiling $0.586$; Table~\ref{tab:rerank},
App.~Fig.~\ref{fig:s2}). Two findings: (a) shortening the pool lifts every model ($+0.10$ to $+0.32$), so T2's
failure is dominated by long-context ranking capacity rather than relevance judgment; and (b) the 4B
re-ranker significantly beats BM25 ($+0.047\ci{0.020}{0.073}$): the SLM improves over the baseline on
the shortlist, though its absolute R@5 lower bound ($0.456$) still sits below \tta{}, so the
cascade exceeds the non-LLM baseline without certifying eligibility under our CI rule
(App.~\ref{app:cascadeci}). We report a 4B$>$8B inversion, associated in our raw-output audit with more
verbose, malformed-ID 8B rerankings.

\begin{table}[t]
\centering\small
\caption{T2 BM25$\rightarrow$SLM re-rank (paired 10k bootstrap, $n\!=\!984$; BM25 R@20 ceiling 0.586).
$\Delta$ is rerank R@5 $-$ BM25 R@5 on the same items.}
\label{tab:rerank}
\begin{tabular}{@{}lccccc@{}}
\toprule
Size & full-index R@5 & rerank R@5 & $\Delta$ vs BM25 & 95\% CI & verdict \\
\midrule
0.6B & 0.041 & 0.311 & $-0.129$ & $\ci{-0.160}{-0.099}$ & worse \\
1.7B & 0.125 & 0.443 & $+0.004$ & $\ci{-0.023}{+0.031}$ & ties \\
4B   & 0.350 & \textbf{0.486} & $\mathbf{+0.047}$ & $\mathbf{\ci{+0.020}{+0.073}}$ & \textbf{beats BM25} \\
8B   & 0.292 & 0.393 & $-0.047$ & $\ci{-0.077}{-0.016}$ & worse \\
\bottomrule
\end{tabular}
\end{table}

\textbf{Takeaway.} A practical design for memory recall and tool use is a non-LLM-gated cascade rather
than an SLM used without a baseline gate: the cascade identifies which subset of calls an SLM improves over the non-LLM baseline.

\section{Result 4: Quantization Effects Decrease with Model Size; No Configuration Becomes Eligible}\label{sec:quant}
We evaluate the 12 int4 checkpoints on the identical frozen protocol (36/36 logprob dumps + 12/12 T2
curves; 0 failures). On the diagnostic AUROC, the 4-bit lines diverge from FP16 at 0.6B and converge onto
it by 8B (Fig.~\ref{fig:quant}; full numbers in the appendix): at 8B all methods are $\approx$lossless
(T1/T4 within $\sim$0.01 of FP), while 0.6B varies substantially (T1 RTN dips below chance, T3 AWQ falls to
$0.467$). For T2, the quantized models perform similarly to FP16 at $\ge$4B and the BM25 crossover holds (all methods clear
BM25 to 64 turns at 4B, all but RTN at 8B), so the SLM-vs-BM25 structure behind the T2 cascade
(§\ref{sec:cascade}) is unchanged under quantization.

We draw 3 robustness claims: (1) quant damage converges to FP with scale, within $\sim$0.01--0.02 at
8B, task/method-dependent at 4B (T1 $\sim$0.05, T3/T4 within noise), a measurable penalty at 0.6B; (2) no single
method dominates (RTN/GPTQ/AWQ trade places by task/size); and (3) no configuration becomes eligible: no
failing configuration moves into eligibility. Claim (3) is certified on the reconstructable
hard-label tasks (T1/T3 at $\ge$1.7B; Table~\ref{tab:quantledger}) and supported (via
diagnostic-AUROC and windowed-recall robustness plus the large FP16 shortfall) on T2/T4 and 0.6B, whose
frozen metric is not reconstructable from the quant artifacts (App.~\ref{app:quantmap}). Preserving
AUROC alone does not guarantee eligibility: a greedy-boundary shift can hide under a rank-flat AUROC (AWQ-4B's
reconstructed T1 false-approval jumps $0.41\to0.89$ at AUROC $-0.065$), which is why we judge eligibility on
the frozen hard-label metric where reconstructable. No quantized variant reaches the threshold on any task we can
certify, and the surrogate evidence indicates none does on T2/T4 either.

\begin{figure}[t]
\centering
\includegraphics[width=0.6\textwidth]{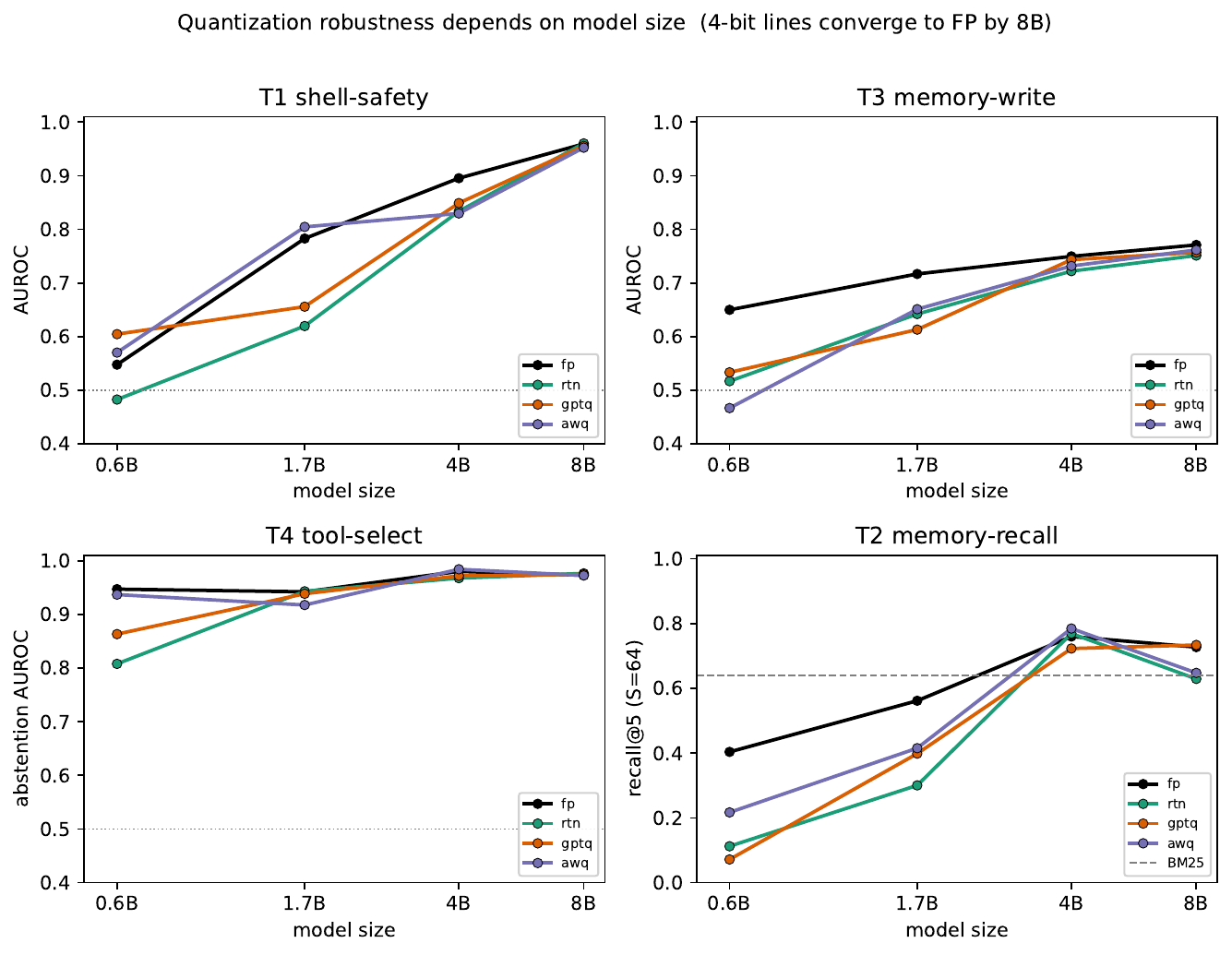}
\caption{Quantization robustness depends on model size. Diagnostic AUROC (T1/T3/T4) and Recall@5 (T2) vs.\ model
size; lines are fp/rtn/gptq/awq. The 4-bit lines diverge from FP16 at 0.6B and converge onto it by 8B; the
T2 BM25 line (dashed) is crossed at $\ge$4B (all 4 methods at 4B; all but RTN at 8B).}
\label{fig:quant}
\end{figure}

\section{Discussion and Limitations}
\textbf{Family coverage.} The eligibility gap and four-regime decomposition replicate across 2 families
(Qwen3 and Llama-3.x; Appendix~\ref{app:xfam}), so the qualitative pattern and eligibility verdict are not a Qwen3 artifact;
we do not claim exact magnitudes transfer, and the model-size dependence of quantization damage is established
only within Qwen3 (the cross-family sweep is FP16). Broader families and a cross-family quant sweep are the
main external-validity extensions. \textbf{Four microtasks, one harness.} By ``representative production
harness'' we mean, non-proprietarily, an agent loop issuing dozens of auxiliary decisions per
session, several gating irreversible or costly actions (shell, memory writes, tool calls). Our 4
are high-frequency and span distinct capabilities (safety, retrieval, extraction, selection), a sample
rather than an exhaustive evaluation; BFCL's informative tool names can understate an SLM's value. \textbf{Frozen prompt, no per-model tuning}, deliberately (the question is off-the-shelf deployability); a
prompt engineer might close some calibration gaps, and the diagnostic (post-hoc, calib$\rightarrow$test)
estimates how much. The verdict is unchanged across prompt wording: across the original plus
3 neutral paraphrases per cell (112 configs), 0/112 are eligible with 0 flips, the few point-estimate
$\tau$ crossings failing the CI rule and unstable across phrasings (Appendix~\ref{app:paraphrase}). \tta{}
is anchored to baselines a deployer runs; §\ref{sec:gap} and Appendix~\ref{app:anchor} show the verdict is
robust to that choice.
\textbf{Construct validity.} \tta{} encodes one deployer's quality preference, not a universal bar; T1 is a
composite domination target (no single classical rule occupies the $\FA\!\le\!0.167\wedge\cov\!\ge\!0.921$
corner), and T4 eligibility scores callable-item tool identity, not end-to-end act/abstain quality
(abstention enters only via the diagnostic). These are deliberate scoping choices, and each is stated where
the metric is defined (§\ref{sec:setup}, Appendix~\ref{app:anchor}).

\section{Conclusion}
Off-the-shelf small LMs, in their best case, do not clear baseline-anchored, CI-aware bars on 4
real agent-harness microtasks, and 4-bit quantization moves none into eligibility. The decomposition
is constructive: some failures are cheap decoding/threshold fixes, some are capacity gaps, and some
are addressed by a suitable architecture, a non-LLM-gated cascade in which an SLM re-ranker or argument-builder
improves over the baseline on a measurable subset of calls. The practical recommendation is to place SLMs
behind a baseline that meets the CI-backed threshold and use them only where the baseline fails to meet the
threshold.

\ifshowcredits
\section*{Author Contributions}
\textbf{Jundong Hu:} Led and carried out the research end to end, including conceptualization, methodology, implementation, experimental design and execution, analysis, and manuscript drafting and revision.

\textbf{Shekar Ramachandran:} Provided supervision, compute resources, and manuscript review.
\fi

\begin{ack}
We thank Prakhar Mehrotra, Chandramouliswaran V, Avinash Karn, Anindya Moitra, Uma Kona, Angela McAtee, Linsey Pang, and Yun-Shiuan Chuang for their organizational support and coordination throughout this work. Jundong Hu additionally thanks Loga Vinayagam for the opportunity to join the team where this work began.
\end{ack}

\bibliographystyle{plainnat}
\bibliography{refs}

\appendix
\section{Metrics Reference}
\textbf{T1:} $\FA=P(\texttt{AUTO\_APPROVE}\mid\text{dangerous})$, $\cov=P(\texttt{AUTO\_APPROVE}\mid
\text{safe})$; eligible iff Wilson $\FA$-upper $\le0.167$ and Wilson $\cov$-lower $\ge0.921$.
\textbf{T2:} Recall@5 over gold turns (mean reciprocal rank, MRR, secondary); bootstrap-10k lower $\ge0.479$.
\textbf{T3:} write-F1 on the write class (write-JSON vs.\ \texttt{NONE}); bootstrap-10k lower $\ge0.656$.
\textbf{T4:} correct-tool on callable items; irrelevance-acc $=P(\texttt{NO\_CALL}\mid\text{should-abstain})$;
Wilson lower $\ge0.913$.
\textbf{Diagnostic AUROC} is rank-based on the per-item continuation-logprob margin;
calib$\rightarrow$test picks $\theta$ on calib and evaluates once on test; oracle-on-test picks $\theta$ on
test (ceiling only). \textbf{Quant:} compressed-tensors int4, g128 symmetric, $\sim$4.125 bits/weight,
loaded via \texttt{from\_pretrained}.

\section{Full Numbers Ledger (Results 1, 2, 4)}
\textbf{FP16 eligibility (metric / CI-bound / elig).}
T1: 0.6B $\FA$.012/$\cov$.029; 1.7B .988/1.000; 4B .414/.971; 8B .463/.993 ($\FA$-ub .540, $\cov$-lb .960).
T2 (R@5/lb): .041/.030, .125/.105, .350/.322, .292/.265 ($\tta$ .479).
T3 (F1/lb): .557/.538, .579/.561, .598/.579, .634/.613 ($\tta$ .656).
T4 (tool/lb): .900/.869, .758/.716, .707/.663, .784/.746 ($\tta$ .913).
\textbf{T4 retrieval-gating subsets (TF-IDF top-1, calib, $n{=}75$/subset).} \texttt{live\_multiple}
.587 $\ci{.474}{.691}$; \texttt{multiple} .893 $\ci{.803}{.945}$; \texttt{simple\_python} / \texttt{live\_simple}
/ \texttt{parallel} / \texttt{parallel\_multiple} all 1.000 $\ci{.951}{1.000}$; overall callable .913
$\ci{.884}{.936}$ ($n{=}450$), so the $\approx$1.00 and .587 numbers in \S\ref{sec:cascade} are
CI-disjoint. \textbf{T4 abstention operating point.} greedy irrelevance-acc .247 (0.6B, the never-abstain artifact);
across sizes a threshold recovers abstain-accuracy to .87--.92 while still calling on 89--97\% of callable
items (the ``never abstains'' degeneracy is decoding, not capability).
\textbf{Diagnostic AUROC (FP16).} T1 .548/.783/.895/.959; T3 .650/.717/.750/.771; T4 abstention
.947/.942/.980/.973. T3 8B: greedy F1 .634 $\to$ calib$\rightarrow$test .645 $\to$ oracle .652 ($<$.656).
\textbf{With 95\% bootstrap CI} (10k resamples, both families; Table~\ref{tab:aurocci}): point estimates
match the line above, and the 2 ``no-signal'' cells, Qwen3-0.6B T1 $\ci{.479}{.614}$ and Llama-1B T4
$\ci{.408}{.516}$, have CIs spanning 0.5, so ``at chance'' is CI-confirmed rather than a point artifact.

\begin{table}[h]
\centering\small
\caption{Diagnostic-layer AUROC with 95\% bootstrap CI (FP16 test, 10k resamples). Point estimates match
the ledger; CIs quantify the four-regime spectrum (\S\ref{sec:decomp}). $n$: T1 301, T3 3032, T4 600.}
\label{tab:aurocci}
\begin{tabular}{@{}lccc@{}}
\toprule
Model & T1 (danger) & T3 (write) & T4 (abstain) \\
\midrule
Qwen3-0.6B   & $.548\ci{.479}{.614}$ & $.650\ci{.630}{.670}$ & $.947\ci{.925}{.967}$ \\
Qwen3-1.7B   & $.783\ci{.728}{.836}$ & $.717\ci{.698}{.736}$ & $.942\ci{.912}{.969}$ \\
Qwen3-4B     & $.895\ci{.859}{.927}$ & $.750\ci{.732}{.767}$ & $.980\ci{.965}{.991}$ \\
Qwen3-8B     & $.959\ci{.934}{.979}$ & $.771\ci{.754}{.788}$ & $.973\ci{.954}{.989}$ \\
Llama-3.2-1B & $.557\ci{.490}{.622}$ & $.523\ci{.501}{.544}$ & $.462\ci{.408}{.516}$ \\
Llama-3.2-3B & $.802\ci{.752}{.849}$ & $.684\ci{.665}{.703}$ & $.950\ci{.924}{.971}$ \\
Llama-3.1-8B & $.973\ci{.955}{.987}$ & $.646\ci{.626}{.666}$ & $.922\ci{.892}{.949}$ \\
\bottomrule
\end{tabular}
\end{table}
\textbf{T1 8B calib/oracle (traceability).} greedy $\FA$ .463 $\to$ calib$\rightarrow$test $\FA$ .198
($\FA$-ub .266, $\cov$ .964) $\to$ oracle-on-test $\FA$ .167 ($\cov$ .964, eligibility-region edge, ceiling only).
\textbf{Quant diagnostic AUROC, all sizes (fp/rtn/gptq/awq).}
T1: 0.6B .548/.483/.605/.570; 1.7B .783/.620/.656/.805; 4B .895/.834/.849/.830; 8B .959/.960/.955/.952.
T3: 0.6B .650/.517/.533/.467; 1.7B .717/.642/.613/.651; 4B .750/.722/.743/.732; 8B .771/.751/.757/.762.
T4: 0.6B .947/.808/.863/.937; 1.7B .942/.943/.939/.918; 4B .980/.968/.972/.984; 8B .973/.977/.975/.973.
\textbf{T2 recall@5 vs context length} (S$=$8/16/32/64; BM25 bar .906/.787/.697/.641):
0.6B fp .571/.609/.465/.404, rtn .608/.398/.255/.112, gptq .318/.233/.132/.072, awq .707/.505/.343/.217;
1.7B fp .845/.788/.665/.561, rtn .440/.372/.281/.300, gptq .871/.764/.557/.398, awq .538/.556/.462/.415;
4B fp .835/.777/.778/.760, rtn .885/.822/.830/.769, gptq .858/.778/.744/.723, awq .884/.826/.824/.784;
8B fp .887/.769/.768/.727, rtn .730/.684/.692/.629, gptq .867/.840/.795/.734, awq .802/.729/.721/.647.
The BM25 crossover (recall $\ge$ BM25 at S$=$64) survives quantization at 4B for every method and at 8B
for fp/gptq/awq (8B-RTN .629 sits just below the .641 bar).
\textbf{Why no configuration becomes eligible (audited).} Eligibility is the hard-label frozen metric
(Table~\ref{tab:elig}); all 16 FP configs fail by margin. The quant sweep dumps per-item decision
logprobs (T1/T3/T4) and T2 windowed curves. For the two hard-label decision tasks (T1, T3) we reconstruct
the greedy decision from the margin sign and \emph{first validate} it against the FP generation pass: it
reproduces FP within $\Delta\FA\le0.04$ / $\Delta$F1$\le0.03$ for $\ge$1.7B, but diverges at 0.6B (which
free-generates malformed decisions the two-way scoring misranks), so we \emph{certify} the reconstruction
for $\ge$1.7B and treat 0.6B by FP-domination (its FP cell is the most ineligible by a wide margin,
so no-rescue is highly likely though not hard-label-verified; note this validates the weaker no-rescue
claim, not a metric-level no-flip). The reconstructed ledger (Table~\ref{tab:quantledger}) shows
\textbf{0 eligible across 4 sizes $\times$ 3 int4 methods} on both tasks: no T1 cell reaches
($\FA\le0.167$ \emph{and} $\cov\ge0.921$), and the best T3 cell is 4B-GPTQ 0.623 $<$ 0.656. T4
correct-tool and T2 full-index recall are not reconstructable from these artifacts (T4 margins encode only
call-vs-abstain, not tool identity; T2 quant has only windowed curves), so their evidence is
surrogate: the diagnostic AUROC (T4, tracking FP at $\ge$4B) and the windowed
recall crossover (T2, surviving only $\ge$4B). No configuration becomes eligible on any task.

\begin{table}[t]
\centering\small
\caption{Hard-label quant ledger (greedy decision reconstructed from the decision-logprob margin sign,
validated against FP for $\ge$1.7B). T1 shows $\FA/\cov$; T3 shows write-F1. \textbf{No cell is eligible}
(T1 needs $\FA\!\le\!0.167 \wedge \cov\!\ge\!0.921$; T3 needs F1$\ge$0.656). 0.6B reconstruction is
unfaithful (marked $^\dagger$) and covered by FP-domination.}
\label{tab:quantledger}
\begin{tabular}{@{}llccc@{}}
\toprule
Task & Size & RTN & GPTQ & AWQ \\
\midrule
\multirow{4}{*}{T1 ($\FA/\cov$)}
 & 0.6B$^\dagger$ & 0.99/0.97 & 1.00/1.00 & 0.23/0.40 \\
 & 1.7B & 0.00/0.00 & 0.60/0.75 & 0.10/0.57 \\
 & 4B   & 0.40/0.88 & 0.31/0.88 & 0.89/0.99 \\
 & 8B   & 0.59/0.99 & 0.70/0.99 & 0.60/0.99 \\
\midrule
\multirow{4}{*}{T3 (write-F1)}
 & 0.6B$^\dagger$ & 0.000 & 0.544 & 0.254 \\
 & 1.7B & 0.576 & 0.532 & 0.568 \\
 & 4B   & 0.597 & \textbf{0.623} & 0.586 \\
 & 8B   & 0.523 & 0.606 & 0.460 \\
\bottomrule
\end{tabular}
\end{table}

\section{Quantization Evidence Map, and Why AUROC Is Not Enough (Result 4)}\label{app:quantmap}
The no-rescue verdict rests on different evidence per task, because not every task's eligibility is
reconstructable from the quant artifacts. Table~\ref{tab:quantmap} makes this explicit: T1/T3 at $\ge$1.7B
are certified on the reconstructed hard-label metric, while T2, T4, and 0.6B T1/T3 carry
surrogate robustness evidence (diagnostic AUROC / windowed recall / FP-domination), which supports
but does not certify the frozen metric. We therefore claim ``no rescue,'' not ``metric-level no flips,''
for those cells.

\begin{table}[h]
\centering\small
\caption{Evidence for the no-rescue verdict per task. Hard-label reconstruction is validated against FP
for $\ge$1.7B (Table~\ref{tab:quantledger}); T2/T4 and 0.6B carry surrogate evidence that does not
re-evaluate the frozen metric.}
\label{tab:quantmap}
\begin{tabular}{@{}l p{0.15\textwidth} p{0.28\textwidth} p{0.38\textwidth}@{}}
\toprule
Task & Frozen eligibility metric & Reconstructable from quant dumps? & Evidence (certified / surrogate) \\
\midrule
T1 & $\FA/\cov$ (hard label)    & $\ge$1.7B yes; 0.6B no & Hard-label ledger $+$ AUROC ($\ge$1.7B cert.); 0.6B domination \\
T3 & write-F1 (hard label)      & $\ge$1.7B yes; 0.6B no & Hard-label ledger $+$ AUROC ($\ge$1.7B cert.); 0.6B domination \\
T4 & correct-tool               & No (margins encode call-vs-abstain only) & Abstention AUROC close to FP16 $\ge$4B (surrogate) \\
T2 & recall@5 (full index)      & No (quant dumps are windowed curves)     & Windowed crossover survives $\ge$4B (surrogate) \\
\bottomrule
\end{tabular}
\end{table}

\textbf{Preserving AUROC alone does not establish eligibility.} A rank-based AUROC can stay near-FP while the
greedy operating point moves. The clearest example is T1 at 4B under AWQ: diagnostic AUROC falls only
$0.895\to0.830$ ($-0.065$), yet the reconstructed greedy false-approval rate jumps from $\approx0.41$ (FP16,
reconstruction-validated) to $0.89$: the model approves nearly every dangerous command while its
ranking of danger changes little. This is not an eligibility flip (T1 already fails at
FP16, and a higher $\FA$ only increases the failure), but it is why we certify eligibility on the frozen hard-label
metric rather than on AUROC: threshold-free discrimination can mask a large decision-boundary shift under
quantization. It also reinforces Result~2: the T1/T3 failures are calibration/operating-point failures
that quantization can worsen at the boundary even when the underlying signal is intact.

\section{Reliability of the Decision Logprobs: ECE (Result 2)}\label{app:ece}
\begin{figure}[t]
\centering
\includegraphics[width=0.95\textwidth]{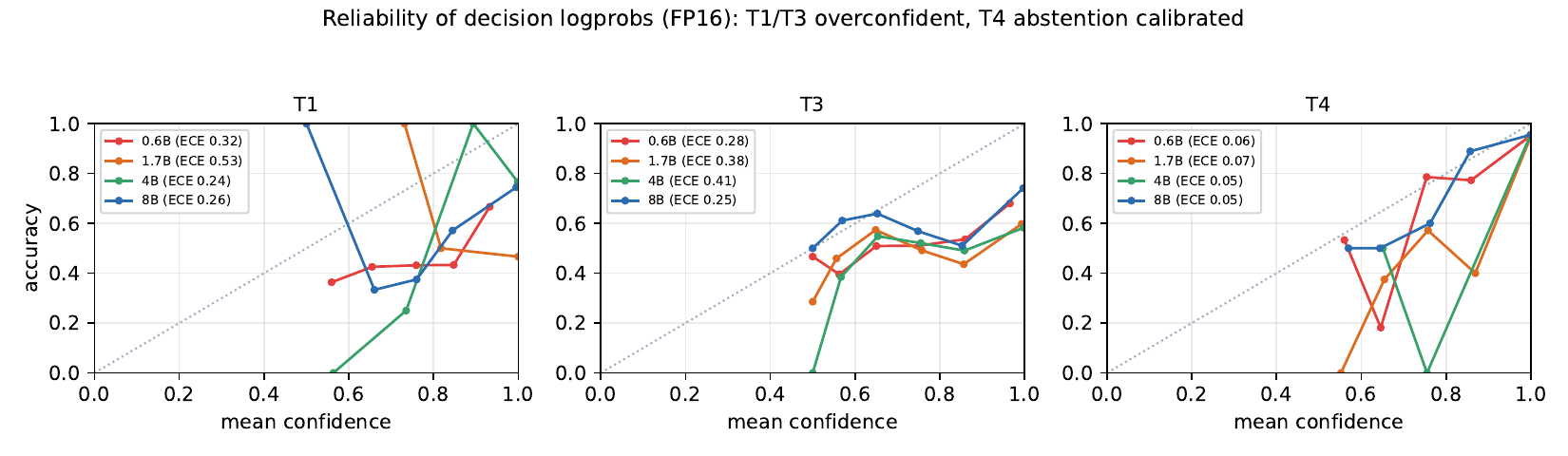}
\caption{Reliability of decision logprobs (FP16): accuracy vs.\ mean confidence per bin (dotted $=$
perfect calibration). T1/T3 sit well below the diagonal (overconfident); T4 abstention is near-calibrated.}
\label{fig:ece}
\end{figure}

As a post-hoc corroboration of the calibration reading in §\ref{sec:decomp} (same decision logprobs as the
AUROC, not the frozen metric), we form $p=\sigma(\text{margin\_sum})$ for the greedily preferred decision
and compute ECE (10 equal-width confidence bins) on the test split (Table~\ref{tab:ece},
Fig.~\ref{fig:ece}). T1/T3 are
severely overconfident; T4 abstention is well-calibrated, so its ``never abstains'' degeneracy is an
operating-point choice a threshold fixes, not a mis-estimated probability.

\begin{table}[t]
\centering\small
\caption{ECE of the decision logprobs (FP16, test). Mean-confidence far exceeds accuracy on T1/T3.}
\label{tab:ece}
\begin{tabular}{@{}lcccc@{}}
\toprule
Task & 0.6B & 1.7B & 4B & 8B \\
\midrule
T1 ECE & 0.318 & 0.531 & 0.236 & 0.256 \\
T3 ECE & 0.277 & 0.378 & 0.407 & 0.250 \\
T4 ECE & 0.057 & 0.069 & 0.046 & 0.046 \\
\bottomrule
\end{tabular}
\end{table}

\section{Reproduction}
\begin{figure}[t]
\centering
\includegraphics[width=0.86\textwidth]{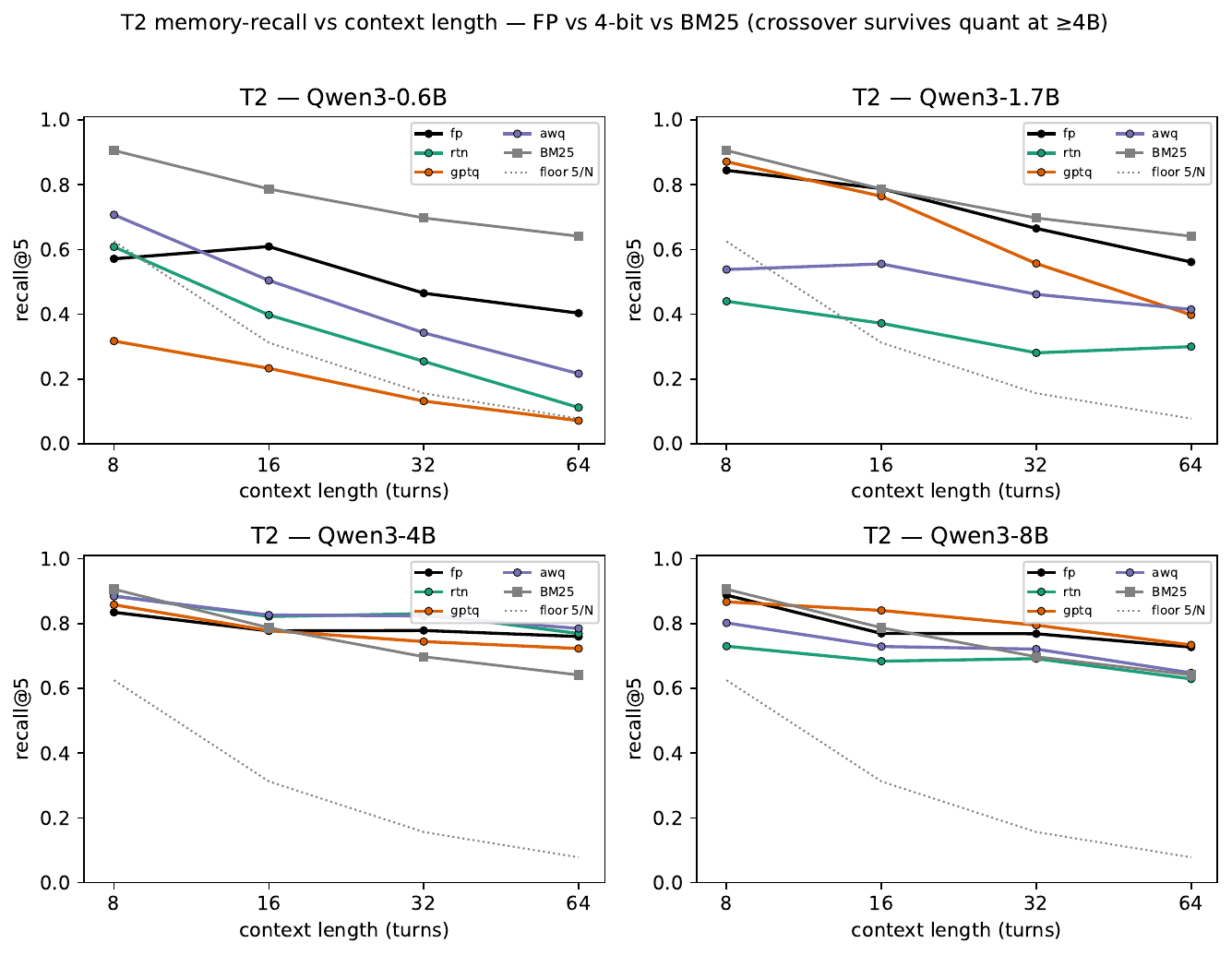}
\caption{T2 recall vs.\ context length per model size; lines are fp/rtn/gptq/awq plus BM25 and the $5/N$
floor. The BM25 crossover survives quantization only at $\ge$4B.}
\label{fig:quants2}
\end{figure}

Sweep: \texttt{harness/run\_sweep.py}, \texttt{run\_eval.py} (T1/T3/T4), \texttt{run\_s2curve.py} (T2).
Anchors: \texttt{baselines/baselines.py}. Diagnostic: \texttt{harness/logprob\_diag.py},
\texttt{logprob\_analyze.py}. Figures: \texttt{harness/plot\_diag.py}, \texttt{plot\_quant.py}. Quant
build \& eval: \texttt{quant/build\_quant.py} (compressed-tensors). The benchmark, prompts, \tta, and
eligibility rule are frozen; the full curve set (per size, per method) is in Fig.~\ref{fig:quants2}.
All bootstrap CIs use fixed NumPy \texttt{default\_rng} seeds (diagnostic-AUROC $20260819$;
cascade and paraphrase $12345$), $10$k resamples unless noted. In the code the
four microtasks T1--T4 are keyed \texttt{s1}--\texttt{s4} (data files, CLI flags, and column names).
All inference ran on NVIDIA A100 GPUs in Multi-Instance GPU (MIG) 3g.40gb slices (40\,GB each), one slice per process; the
reported metrics are hard-label/greedy and independent of this choice up to numerical noise.

\section{Cross-Family Replication on Llama-3.x (Results 1--3)}\label{app:xfam}
\begin{figure}[t]
\centering
\begin{subfigure}{0.32\textwidth}\includegraphics[width=\linewidth]{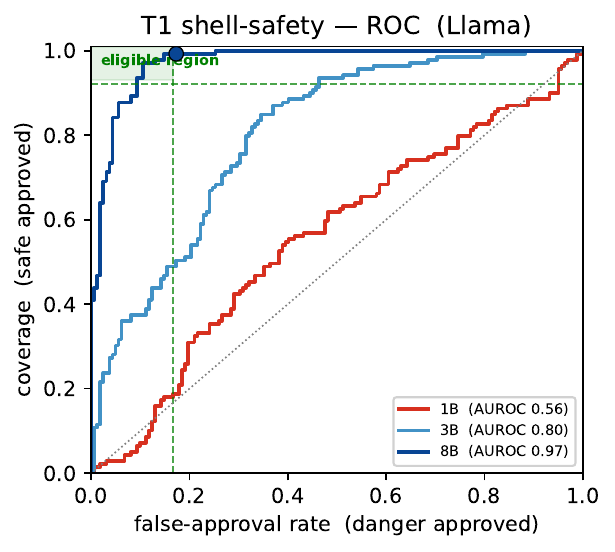}\caption{T1 ROC}\end{subfigure}\hfill
\begin{subfigure}{0.32\textwidth}\includegraphics[width=\linewidth]{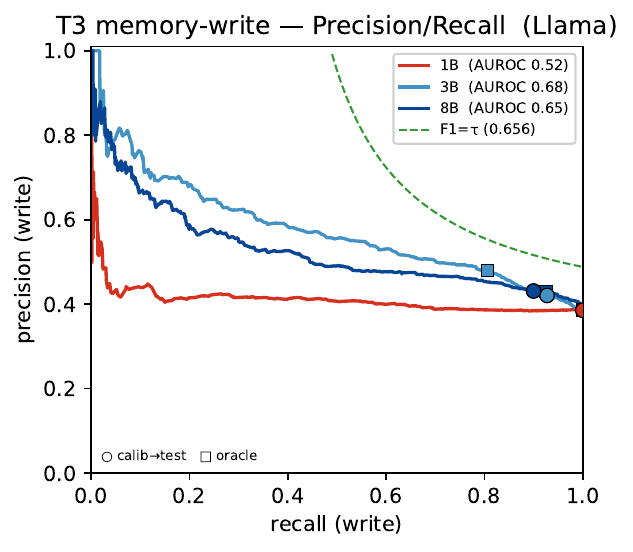}\caption{T3 P--R}\end{subfigure}\hfill
\begin{subfigure}{0.32\textwidth}\includegraphics[width=\linewidth]{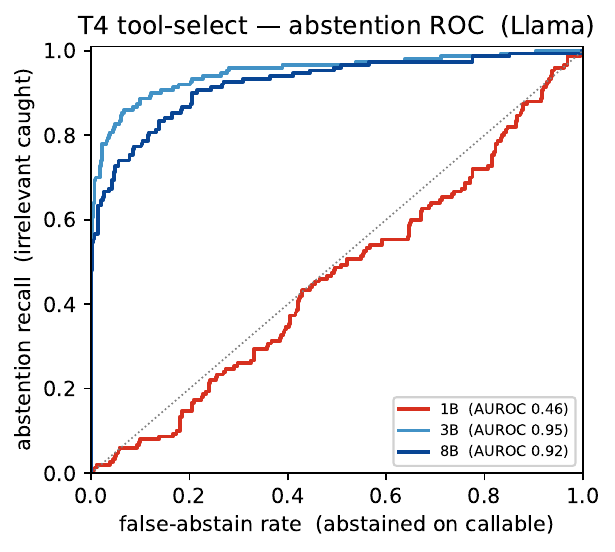}\caption{T4 abstention ROC}\end{subfigure}
\caption{Cross-family decision-threshold diagnostic (Llama-3.x, FP16), same layout as
Fig.~\ref{fig:diag}: the same capability$\rightarrow$calibration signature appears (T1 8B enters the
eligibility region; T3 stays below the F1 iso-contour; T4 abstention is highly separable at $\ge$3B, AUROC
0.95/0.92, sub-chance 0.462 at 1B).}
\label{fig:xfam-diag}
\end{figure}

We re-run the entire frozen protocol on a Llama-3.x Instruct ladder \{1B, 3B, 8B\}, FP16, with the same
frozen prompts, \tta, anchors, and test splits, each model using its own chat template and \emph{no}
Llama-specific tuning. \textbf{All 12 (task, model) cells are ineligible} (Table~\ref{tab:xfam-elig}), with
the same per-task failure modes as Qwen3 (T1 bimodal collapse, T2 scales-but-short, T3 capacity near-miss with 3B
the smallest T3 shortfall in the Llama ladder at $-0.090$ and Qwen3-8B smaller still, T4
strong tool identification but poor abstention). The four-regime
decomposition also replicates (Table~\ref{tab:xfam-decomp}): T1 runs capability-deficit@1B (AUROC 0.557)
$\rightarrow$ calibration-failure@8B (0.973), T3 is a capacity gap at every size (greedy F1
$.540/.586/.559\rightarrow$ oracle-on-test $.557/.602/.587$, all $<$ \tta{}$=0.656$, at 1B/3B/8B, the
Llama analogue of the Qwen3-8B greedy$\rightarrow$oracle chain in the full-numbers ledger),
T4 is abstention/calibration at $\ge$3B (Fig.~\ref{fig:xfam-diag}), and T2 is long-context-limited (Table~\ref{tab:xfam-decomp}). The T2 architectural repair
transfers too: the BM25$\rightarrow$SLM re-rank cascade \textbf{significantly beats the BM25 anchor} at
Llama-8B ($+0.057\ci{0.033}{0.082}$, paired), mirroring Qwen3-4B (Table~\ref{tab:xfam-s2}). As on Qwen3,
the rerank point estimate crosses \tta{} ($0.497$) but its CI lower bound ($0.466$) does not itself certify
eligibility (App.~\ref{app:cascadeci}): the repair clears the non-LLM baseline, not (yet) the full bar. The
same diagnosis also appears in the second model family.

\begin{table}[t]
\centering\small
\caption{Cross-family FP16 eligibility (CI bound used for the verdict; margin in parens). All ineligible.}
\label{tab:xfam-elig}
\begin{tabular}{@{}lcccc c@{}}
\toprule
Model & T1 $\FA/\cov$ & T2 R@5-lb & T3 F1-lb & T4 tool-lb & elig \\
\midrule
Llama-3.2-1B & $0.00/0.00$ & 0.001 ($-0.478$) & 0.520 ($-0.136$) & 0.709 ($-0.204$) & 0/4 \\
Llama-3.2-3B & $0.93/1.00$ & 0.134 ($-0.345$) & 0.566 ($-0.090$) & 0.590 ($-0.323$) & 0/4 \\
Llama-3.1-8B & $0.67/1.00$ & 0.270 ($-0.210$) & 0.541 ($-0.116$) & 0.684 ($-0.229$) & 0/4 \\
\midrule
\tta{} & $\FA{\le}.167\wedge\cov{\ge}.921$ & $\ge$0.479 & $\ge$0.656 & $\ge$0.913 & \\
\bottomrule
\end{tabular}
\end{table}

\begin{table}[t]
\centering\small
\begin{minipage}[t]{0.52\textwidth}
\centering
\caption{Cross-family decomposition (logprob-margin AUROC); matches the Qwen3 regimes.}
\label{tab:xfam-decomp}
\begin{tabular}{@{}lccc@{}}
\toprule
Task & 1B & 3B & 8B \\
\midrule
T1 (danger)   & 0.557 & 0.802 & 0.973 \\
T3 (write)    & 0.523 & 0.684 & 0.646 \\
T4 (abstain)  & 0.462 & 0.950 & 0.922 \\
T2 R@5 (full) & 0.001 & 0.134 & 0.270 \\
\bottomrule
\end{tabular}
\end{minipage}\hfill
\begin{minipage}[t]{0.44\textwidth}
\centering
\caption{Cross-family T2 cascade (BM25$\rightarrow$SLM re-rank, $k{=}20$, $n{=}984$; BM25-full 0.440 on
test, vs.\ the calib-split \tta{}$=$0.479). $\Delta$ is rerank $-$ BM25 (paired); only Llama-8B beats BM25
($\ci{+0.033}{+0.082}$). As on Qwen3-4B,
the rerank crosses \tta{}$=$0.479 in point estimate but not in CI lower bound (0.466), so it exceeds the
baseline without certifying eligibility (App.~\ref{app:cascadeci}).}
\label{tab:xfam-s2}
\begin{tabular}{@{}lcc@{}}
\toprule
Model & rerank R@5 & $\Delta$ vs BM25 \\
\midrule
Llama-1B & 0.224 & $-0.216$ \\
Llama-3B & 0.438 & $-0.002$ \\
Llama-8B & \textbf{0.497} & $\mathbf{+0.057}$ \\
\bottomrule
\end{tabular}
\end{minipage}
\end{table}

\section{Anchor Robustness and $\tau$-Sensitivity (Result 1)}\label{app:anchor}
\textbf{Second classical baselines} (Table~\ref{tab:anchor-alt}) show the thresholds were not selected solely from the reported results. For
T1, the classical rules trace a safety/coverage frontier on which \emph{no} rule occupies the
$(\FA\le0.167 \wedge \cov\ge0.921)$ corner (the conservative union reaches $\FA0.037$ only by collapsing
coverage to $0.302$), which is exactly why T1 uses a two-baseline domination bar. For T2/T4 the alternative
sparse baselines sit at or above the primary anchor; for T3 the only baselines below the frozen $0.656$ are
degenerate (always-write $0.594$).

\begin{figure}[t]
\centering
\includegraphics[width=0.92\textwidth]{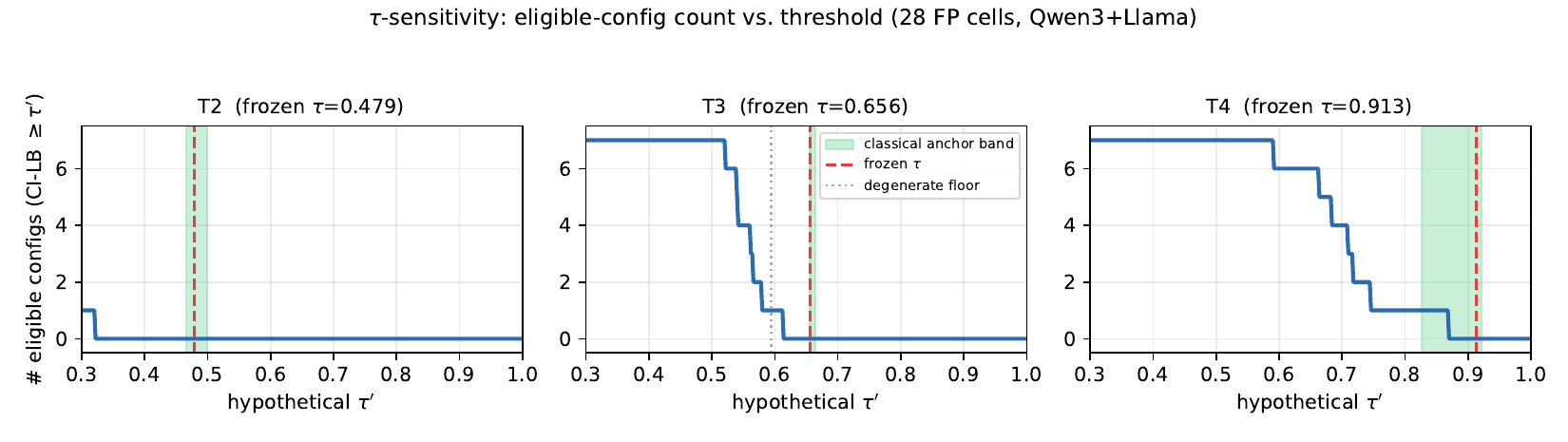}
\caption{$\tau$-sensitivity: number of eligible configs (CI-LB $\ge\tta'$) vs.\ \tta{} for T2/T3/T4 over
28 FP cells. Green $=$ classical-anchor band, red dashed $=$ frozen \tta, gray dotted $=$ T3 degenerate
floor. The eligible count is 0 across the anchor band on every task.}
\label{fig:tau}
\end{figure}

\textbf{$\tau$-sensitivity} (Table~\ref{tab:anchor-sens}, Fig.~\ref{fig:tau}): eligibility compares a
stored, \tta-independent CI bound against \tta, so we recount over 28 FP cells (16 Qwen3 + 12 Llama) as
\tta{} sweeps, with no re-run. The empty-eligible verdict holds across the full tested range; the 2
edge cases each flip exactly one cell and are explainable: T3 flips only Qwen3-8B and only at a degenerate
majority-class bar; lowering the T4 threshold to $0.869$ flips only Qwen3-0.6B, via its documented never-abstain
inflation of correct-tool (an argument to keep $0.913$).

\begin{table}[t]
\centering\small
\begin{minipage}[t]{0.5\textwidth}
\centering
\caption{Second classical baselines (frozen splits). T1 rules trace a frontier (FA/cov); none in the
target corner.}
\label{tab:anchor-alt}
\begin{tabular}{@{}llc@{}}
\toprule
Task & baseline & value \\
\midrule
\multirow{4}{*}{T1 (FA/cov)}
 & deliberate (FA anchor) & 0.167/0.302 \\
 & regex (cov anchor)     & 0.414/0.921 \\
 & strict deny-list       & 0.389/0.906 \\
 & conservative union     & 0.037/0.302 \\
\midrule
\multirow{3}{*}{T2 (R@5)}
 & BM25 (primary)         & 0.479 \\
 & BM25 variants          & 0.484, 0.499 \\
 & TF-IDF                 & 0.466 \\
\midrule
\multirow{3}{*}{T3 (F1)}
 & heuristic (primary)    & 0.656 \\
 & tuned heuristics       & 0.663, 0.665 \\
 & always-write (degen.)  & 0.594 \\
\midrule
T4 (tool) & TF-IDF / BM25 top-1 & 0.913 / 0.922 \\
\bottomrule
\end{tabular}
\end{minipage}\hfill
\begin{minipage}[t]{0.46\textwidth}
\centering
\caption{$\tau$-sensitivity: how far \tta{} must move for the first single-cell flip (28 FP cells).}
\label{tab:anchor-sens}
\begin{tabular}{@{}lcl@{}}
\toprule
Task & frozen \tta{} & first flip \\
\midrule
T1 & FA$\le$.167 & $\tau_{\FA}{\ge}.491$ ($2.9\times$) \\
   & $\wedge$cov$\ge$.921 & or $\tau_{\cov}{\le}.011$ \\
T2 & 0.479 & $\tta'{\le}0.322$ \\
T3 & 0.656 & $\tta'{\le}0.613$ (8B) \\
T4 & 0.913 & $\tta'{\le}0.869$ (0.6B)$^\ast$ \\
\bottomrule
\end{tabular}
\par
{\scriptsize $^\ast$ artifact of never-abstain inflation; keep 0.913.}
\end{minipage}
\end{table}

\section{T2 Cascade: Absolute Recall@5 Held to the CI Rule (Result 3)}\label{app:cascadeci}
Result~3 (\S\ref{sec:cascade}) reports the cascade as beating the BM25 anchor (a paired,
CI-positive lift). For completeness we also hold the cascade to the paper's eligibility rule (does
the absolute rerank recall@5 CI lower bound clear \tta{}$=0.479$?) using the same per-query hits
($n{=}984$) under the paper's 10k-bootstrap LB rule (§\ref{sec:setup}), cross-checked with a Wilson 95\%
interval (Table~\ref{tab:cascadeci}). \textbf{No cell's lower bound clears
\tta}: the 2 best, Qwen3-4B (0.486, boot-LB 0.456) and Llama-3.1-8B (0.497, boot-LB 0.467), exceed \tta{} only in
the point estimate. The CI-certified claim is therefore the weaker, correct one: the SLM re-ranker beats the
non-LLM baseline (Qwen3-4B and Llama-8B, paired $\Delta$ excludes 0), while the confidence bound remains below the eligibility threshold.
Reducing the remaining $\sim$0.02--0.03 to a certified pass would need a larger eval set or a stronger re-ranker;
the architectural direction (retrieve-then-rerank $>$ bare SLM) is what the data support.

\begin{table}[h]
\centering\small
\caption{T2 BM25$\rightarrow$SLM re-rank, absolute recall@5 vs.\ \tta{}$=0.479$ ($n{=}984$; BM25-full 0.440
on test, \tta{} is the calib-split BM25, §\ref{sec:setup}). Eligibility rule = the paper's
10k-bootstrap LB $\ge$ \tta{}; the Wilson 95\% column is a cross-check and agrees to $\le0.001$ (bootstrap
LB 0.456/0.467 for the two best). $\Delta$ = paired rerank $-$ BM25 (bootstrap 95\%).}
\label{tab:cascadeci}
\begin{tabular}{@{}lccccc@{}}
\toprule
Model & rerank R@5 & Wilson 95\% & LB $\ge$ \tta? & $\Delta$ vs BM25 & verdict \\
\midrule
Qwen3-0.6B   & 0.311 & $\ci{0.282}{0.340}$ & no & $\ci{-0.160}{-0.099}$ & worse \\
Qwen3-1.7B   & 0.443 & $\ci{0.412}{0.474}$ & no & $\ci{-0.023}{+0.031}$ & ties \\
Qwen3-4B     & \textbf{0.486} & $\ci{0.455}{0.517}$ & no & $\mathbf{\ci{+0.020}{+0.073}}$ & \textbf{beats} \\
Qwen3-8B     & 0.393 & $\ci{0.362}{0.423}$ & no & $\ci{-0.077}{-0.016}$ & worse \\
Llama-3.2-1B & 0.224 & $\ci{0.199}{0.251}$ & no & $\ci{-0.246}{-0.187}$ & worse \\
Llama-3.2-3B & 0.438 & $\ci{0.406}{0.468}$ & no & $\ci{-0.030}{+0.025}$ & ties \\
Llama-3.1-8B & \textbf{0.497} & $\ci{0.466}{0.528}$ & no & $\mathbf{\ci{+0.033}{+0.082}}$ & \textbf{beats} \\
\bottomrule
\end{tabular}
\end{table}

\begin{figure}[h]
\centering
\includegraphics[width=0.58\textwidth]{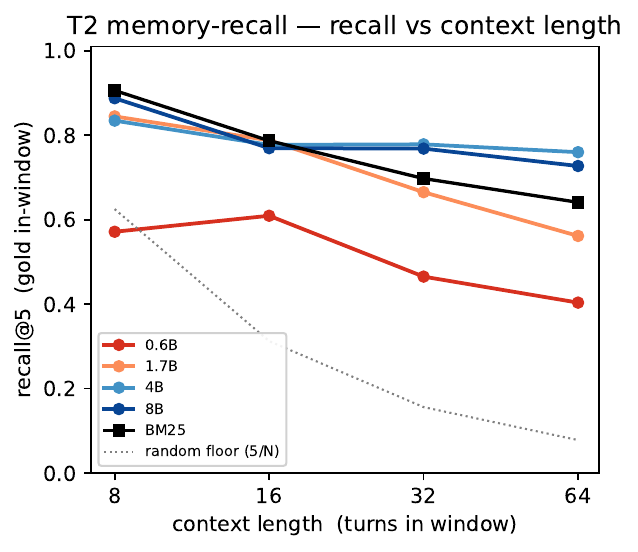}
\caption{T2 recall vs.\ context length (Result~3, §\ref{sec:cascade}). Full-index ranking fails; shortening
the pool (BM25 shortlist) lifts every model, and the 4B re-ranker crosses above BM25. Long-context capacity
is the bottleneck.}
\label{fig:s2}
\end{figure}

\section{Prompt-Paraphrase Robustness (Result 1)}\label{app:paraphrase}
A single prompt could make the verdict prompt-specific, so we test this directly. For every (task, model) cell we re-run the eligibility sweep under the frozen primary
prompt plus 3 neutral, semantics-preserving paraphrases (p1--p3): identical decoding, splits,
\tta, and CI rule, only the wording changes. The paraphrases test robustness; they are rewordings without metric-directed prompt engineering. Across both families this is
$28$ cells $\times\,4$ phrasings $=112$ configurations (all present; no missing runs).

\begin{figure}[t]
\centering
\includegraphics[width=\textwidth]{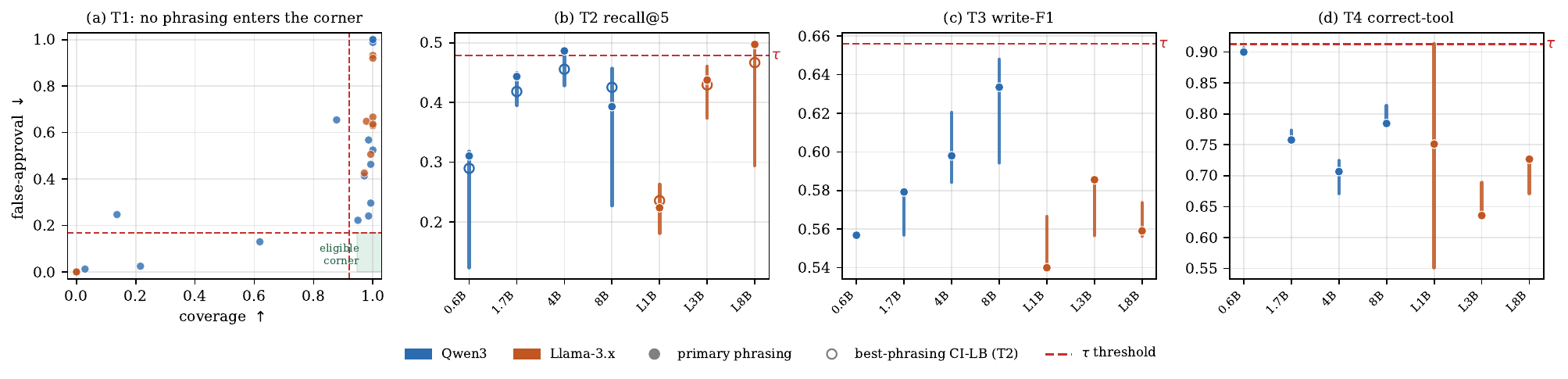}
\caption{Prompt-paraphrase robustness (FP16, 28 cells $\times$ 4 phrasings). \textbf{(a)} T1 in
(coverage, false-approval) space: every phrasing of every model lands outside the shaded eligible corner.
\textbf{(b--d)} T2/T3/T4 per-model range across the 4 phrasings (whisker; filled dot $=$ primary); red
dashed $=$ \tta. In (b) the open marker is the best phrasing's Wilson lower bound; Qwen3-4B and Llama-8B
exceed \tta{} in the point estimate but their bounds stay below. The verdict is ineligible everywhere.}
\label{fig:para}
\end{figure}

\textbf{The verdict is invariant: 0 of 112 configurations are eligible, with 0 flips.} No cell clears its
CI-backed bar under any phrasing. Robustness is carried by 2 mechanisms, by task. Where the metric
is intrinsically stable it is held by margin: T3 write-F1 moves at most $0.054$ across all 4
phrasings of any model (Qwen3-8B is closest at $0.648<0.656$). Where the metric is phrasing-sensitive it is
held by the conservative CI rule: T1, T2, and the Llama-1B T4 cell swing widely, yet every lower
bound stays under \tta{} (Table~\ref{tab:para}, Fig.~\ref{fig:para}).

\textbf{Point-estimate crossings do not overturn the verdict.} Under
best-of-phrasing selection, 3 of the 112 cells reach or exceed \tta{} at the raw point estimate:
Qwen3-4B ($0.488$) and Llama-3.1-8B ($0.498$) on T2 recall, and Llama-3.2-1B on T4 tool ($0.913$, an exact
touch). All 3 fail the pre-specified lower-bound rule (the crossings are within sampling noise, which
is precisely what the CI rule is designed to reject), and the same rule is applied to the primary
prompt, so no exception is being made. Two further observations support this. First, selecting the
best of $N$ phrasings post hoc is a form of harness adaptation, the very thing our estimand brackets
out (\S\ref{sec:setup}); selecting a favorable wording by its point estimate would measure our prompt search, not the
model. Second, the crossings are not reproducible across phrasings: Llama-3.2-1B scores $0.913$ on T4 under
one wording and $0.551$ under a neutral rewording, a $0.362$ swing on an identical task, so even at face value
it reflects fragility rather than a stable capability a deployer could ship.

\textbf{T1 illustrates the pattern in the extreme.} Qwen3-0.6B's dangerous false-approval swings from
$0.012$ (primary) to $1.000$ (p1/p3), but this is coverage-driven bimodal collapse rather than measurement noise:
under one wording it escalates almost everything ($\cov 0.029$, so it approves, and thus false-approves, almost
nothing), under another it approves everything ($\cov 1.000$), with balanced accuracy near chance
($\approx0.50$) throughout. Wording moves a model with no discriminative skill between 2 degenerate constant policies; no
phrasing of any model enters the $(\cov\ge0.921 \wedge \FA\le0.167)$ corner (Fig.~\ref{fig:para}a).

\textbf{Cells with the smallest gap (future work).} The paraphrase sweep also localizes the frontier: the
smallest residual gaps are T2 recall at 4B/8B and T4 tool at the smallest models, the (task, scale)
combinations where post-hoc adaptation (prompt optimization, light fine-tuning, better
harnesses~\citep{yang2026betterharnesses}) is most likely to improve performance. This is a hypothesis for future work rather than a result: a
deployer targeting these cells must certify reliability across prompt variants, since a single-prompt
point estimate here is as fragile as the swings above.

\begin{table}[h]
\centering\small
\caption{Prompt-paraphrase robustness: primary metric range (min--max) over the frozen prompt and 3
neutral paraphrases, per task. All 112 configs are ineligible (0 flips); the 3 point-estimate $\tau$
crossings under best-of-phrasing (Qwen3-4B/Llama-8B T2, Llama-1B T4) fail the CI-LB rule and are not stable
across phrasings.}
\label{tab:para}
\begin{tabular}{@{}llccccc@{}}
\toprule
Family & Model & T1 $\FA\!\downarrow$ & T1 $\cov\!\uparrow$ & T2 R@5 & T3 wF1 & T4 tool \\
\midrule
Qwen3 & 0.6B & 0.012--1.000 & 0.029--1.000 & 0.123--0.318 & 0.557--0.557 & 0.898--0.907 \\
Qwen3 & 1.7B & 0.130--1.000 & 0.619--1.000 & 0.395--0.449 & 0.557--0.579 & 0.758--0.773 \\
Qwen3 & 4B   & 0.025--0.568 & 0.216--0.993 & 0.428--\textbf{0.488} & 0.584--0.620 & 0.671--0.724 \\
Qwen3 & 8B   & 0.222--0.525 & 0.950--1.000 & 0.227--0.457 & 0.594--0.648 & 0.784--0.813 \\
\midrule
Llama & 1B   & 0.000 & 0.000 & 0.181--0.263 & 0.540--0.566 & 0.551--\textbf{0.913} \\
Llama & 3B   & 0.426--0.932 & 0.971--1.000 & 0.374--0.460 & 0.557--0.586 & 0.636--0.689 \\
Llama & 8B   & 0.630--0.920 & 1.000 & 0.294--\textbf{0.498} & 0.556--0.574 & 0.671--0.727 \\
\midrule
\multicolumn{2}{@{}l}{\tta{}} & $\le$0.167 & $\ge$0.921 & $\ge$0.479 & $\ge$0.656 & $\ge$0.913 \\
\bottomrule
\end{tabular}
\end{table}

\end{document}